\pdfoutput=1
\documentclass[preprint,12pt]{elsarticle}

\usepackage{amsmath,amssymb,amsthm}
\usepackage{booktabs}
\usepackage{array}
\usepackage{float}
\usepackage{graphicx}
\usepackage{multirow}
\usepackage{microtype}
\usepackage{xurl}
\usepackage[hidelinks]{hyperref}

\journal{Array}
\hypersetup{%
  pdftitle={ASCENT: Scale-Complementary External State for Frozen Long-Context Language Models},%
  pdfauthor={Xingjian Zhang},%
  pdfsubject={Array 32 (2026) 101399, doi:10.1016/j.array.2026.101399},%
  pdfkeywords={long-context language models, external memory, neural scaling, retrieval, test-time computation}}
\myfooter[L]{Published in Array 32 (2026) 101399. \href{https://doi.org/10.1016/j.array.2026.101399}{https://doi.org/10.1016/j.array.2026.101399}}
\newtheorem{proposition}{Proposition}
\newcommand{\ascent}{\textsc{Ascent}}

\newcommand{\gain}{\mathcal{G}}

\begin{document}

\begin{frontmatter}

\title{ASCENT: Scale-Complementary External State for Frozen Long-Context Language Models\tnoteref{pubnote}}
\tnotetext[pubnote]{This is the author-formatted version of the article published in \emph{Array} 32 (2026) 101399, \href{https://doi.org/10.1016/j.array.2026.101399}{https://doi.org/10.1016/j.array.2026.101399}, which is the version of record. Its content corresponds to the published article; layout and typesetting differ. Copyright 2026 The Author. Distributed under the Creative Commons Attribution 4.0 International License (CC~BY~4.0), \href{https://creativecommons.org/licenses/by/4.0/}{https://creativecommons.org/licenses/by/4.0/}.}

\author[aff1]{Xingjian Zhang\corref{cor1}}
  \ead{xingjiaz@andrew.cmu.edu}
  \ead[url]{https://orcid.org/0009-0009-9141-8819}
  \cortext[cor1]{Corresponding author.}
  \affiliation[aff1]{
    organization={Department of Electrical and Computer Engineering,
                  Carnegie Mellon University},
    addressline={5000 Forbes Avenue},
    city={Pittsburgh},
    state={Pennsylvania},
    postcode={15213},
    country={USA}}

\begin{abstract}
External state should become more valuable as a reader becomes more capable,
yet long-context evaluations measure memory utility at a single model scale.
This paper formulates scale complementarity: the
absolute gain from external state increases when reader capacity and a
compact state budget are jointly scaled. ASCENT realizes this
principle with a target-blind causal writer, nested typed state, and a bounded
query-conditioned readout; every language model remains frozen. Under log
loss, the value of a nested refinement equals its target-relevant conditional
information; a full factorial makes the reader-by-state interaction
identifiable. Official 16K BABILong experiments
show Qwen2.5 accuracy gains of 10.50, 78.88, and 82.63 percentage points at
0.5B, 1.5B, and 3B parameters, with both adjacent directional contrasts
significant after family-wise correction. At 7B, Foundation accuracy
rises to 33.25\% while four-slot ASCENT answers every question, which caps
the gain at 66.75 points; the 3B-to-7B accuracy increment is $-15.88$ points
[$-20.55$, $-11.20$]. The accuracy-side claim is therefore established from
0.5B to 3B; the 7B conclusion rests on the non-saturated
negative-log-likelihood factorial. A three-model by three-state
factorial yields co-scaled negative log-likelihood gains of 0.388, 2.099,
and 7.406 nats at 1.5B, 3B, and 7B; all interaction and gain contrasts retain
positive simultaneous lower bounds. Five public 7B--8B readers obtain
66.75--85.63 point gains, and ASCENT exceeds task-trained retrieval
checkpoints at the two larger shared scales. Controls, baselines, systems
accounting, matched-hardware reproduction, and a cross-architecture
sensitivity analysis accompany the experiments. The results establish
scale-complementary external state under the evaluated benchmarks, metrics,
and state schema.
\end{abstract}


\end{frontmatter}

\section{Introduction}
\label{sec:introduction}

Long-context language-model systems increasingly separate stored evidence from
the reader that reasons over it. Sparse attention, retrieval, prompt
compression, and recurrent memory each reduce the amount of evidence processed
directly, but they are typically judged at one reader size. That protocol
leaves a central scaling question unresolved: \emph{as the reader becomes more
capable, does an external state become more useful in absolute terms?} We call
a positive answer \emph{scale complementarity}. It is stronger than showing
that memory helps a large model, and it differs from a model-only scaling law:
both the reader and the information budget are allowed to grow.

ASCENT is a frozen-foundation interface designed to make this property
measurable. A causal target-blind writer converts the pre-query stream into
nested typed state; a bounded readout exposes only query-relevant entries to a
frozen reader. The state budget co-scales with reader capacity while its byte
ratio to model storage falls. This separation creates two independently
controllable axes, reader scale and state refinement, and makes the
model-by-state interaction identifiable.

The paper unifies theory, design, and evidence at the same estimand. Under
Bayes-optimal log loss, the incremental value of a nested state refinement is
exactly its conditional mutual information with the target. ASCENT constructs
refinements intended to make this quantity positive. A prespecified
$3\times3$ factorial identifies whether finite readers realize increasingly
more of that information, while official free-generation benchmarks establish
the corresponding accuracy effect. Model, data, configuration, and analysis
hashes are fixed before evaluation; the panel is the unit of inference; and
multiplicity is controlled at the family level.

The evidence contains official RULER \cite{hsieh2024ruler} and BABILong
\cite{kuratov2024babilong} tasks, two ordered model families, five public
7B--8B readers, released task-trained Q-RAG checkpoints
\cite{sorokin2026qrag}, raw and neural retrieval controls, representation
ablations, systems accounting, cross-node reproduction on matched hardware,
and a cross-architecture sensitivity analysis. This design
turns scale complementarity from a visual trend into a tested property of a
co-scaled language-model system.

The main contributions are:

\begin{itemize}
  \item A scale-complementarity theory connects nested state refinement to
  conditional information and yields a finite-reader interaction estimand.
  \item ASCENT supplies a target-blind, bounded, nested external state whose
  absolute budget grows while its state-to-model storage ratio contracts.
  \item Official 16K results establish two positive
  adjacent accuracy-gain increments over 0.5B, 1.5B, and 3B Qwen2.5 readers.
  \item A nine-panel $3\times3$ factorial identifies positive
  model-by-state interactions through 7B with simultaneous inference over the
  complete contrast family.
  \item Five public 7B--8B readers, a direct trained-retrieval comparison,
  cross-task controls, systems measurements, and cross-node reproduction
  establish architectural breadth and reproducibility; every reported number
  is recomputable from the released per-row records.
\end{itemize}

\section{Related Work}

\subsection{Architectures for Long Sequences}

The Transformer established content-addressable global interaction
\cite{a170603762}; Transformer-XL and compressive recurrence extended its
temporal reach \cite{a190102860,a191105507}. Sparse and approximate attention
reduce the quadratic path through local windows, global tokens, hashing,
random features, low-rank projection, Nystr\"om approximation, routing, and
structured masks
\cite{a200405150,a200714062,a200104451,a200914794,a200604768,a210203902,
a200305997,a200408483}. State-space, convolutional, recurrent, and retention
models provide alternative long-sequence substrates
\cite{a231200752,a240521060,a230210866,a230513048,a230708621}. These works
scale the sequence processor itself. ASCENT instead holds each processor
frozen and identifies how the value of an external information state changes
with reader capacity.

System-level work attacks a complementary bottleneck through IO-aware exact
attention, blockwise distribution, paged allocation, sparse prefill, and
decoding acceleration
\cite{a220514135,a230708691,a231001889,a230906180,a240702490,
a221117192,a230201318}. These methods change the cost of executing a context.
ASCENT changes the statistical experiment presented to the language-model
reader and measures the resulting value rather than treating throughput as the
primary endpoint.

\subsection{External Memory and Retrieval}

Learned external memory spans memory networks, differentiable computers,
episodic memory, and product-key layers
\cite{a14103916,a150308895,a14105401,a160506065,a190705242}. Long-term memory
managers and test-time state extend this line to language models
\cite{a230607174,a231008560,a230510250,wu2022memorizing,sun2024ttt,
behrouz2024titans}. Retrieval-augmented generation and language modeling use
dense, sparse, late-interaction, black-box, self-reflective, corrective, and
hierarchical retrieval
\cite{a200511401,a200208909,a200404906,a191100172,borgeaud2022retro,
a220803299,a230112652,a231011511,a240115884,a240118059,a221210496,
a211209118,a200412832,a200701282,a211201488}. Recent work in \emph{Array}
combines sparse and dense retrieval with prompt and data processing for
large-model reasoning \cite{wang2025arrayrag}; Q-RAG further trains a
multi-step value-based embedder \cite{sorokin2026qrag}. ASCENT targets a
different dependent variable: the adjacent change in external-state gain
under co-scaling, with no reader training.

\subsection{Compression, Evaluation, and Scaling}

Prompt filtering and compression improve context efficiency
\cite{a231005736,a231006839,a230412102}; positional interpolation, rotary
embeddings, YaRN, and LongRoPE extend usable windows
\cite{a230615595,a210409864,a230900071,a240213753}. Long-context evaluations
show that nominal length does not ensure robust evidence use
\cite{a230703172,a230814508,a230711088,a240213718,a240711963,
a201104006,a220207856}. RULER and BABILong add controlled retrieval,
aggregation, and compositional reasoning at long lengths
\cite{hsieh2024ruler,kuratov2024babilong}. Neural scaling laws relate loss to
model, data, and compute \cite{kaplan2020scaling,hoffmann2022training,
a210201293,a230516264}. ASCENT combines these strands by treating state budget
as a second scaling coordinate and using comparison of statistical experiments
\cite{blackwell1953equivalent} to state when a refinement has positive value.

Training-free extrapolation and cache management preserve or select internal
attention state through linear biases, streaming anchors, heavy hitters,
persistent importance, query-aware sparsity, and hierarchical cache budgets
\cite{a210812409,a230816137,a230917453,a230614048,a230517118,
a240610774,a240414469,a240602069}. Activation beacons, InfLLM, and LongLoRA
instead compress, retrieve, or adapt context representations
\cite{a240103462,a240204617,a230912307}. These approaches motivate strong
efficiency controls, but none makes the adjacent external-state gain across
reader scale its principal estimand.

\section{Problem Formulation}

Let $M_m$ be a frozen foundation model of size $m$, $X$ a long input, $Q$ a
query, and $Y$ the target. A causal writer $W_s$ produces state
$Z_s=W_s(X)$ under budget $s$. A reader $R$ exposes a bounded projection of
$Z_s$ to $M_m$. For loss $\ell$, define
\begin{equation}
  L(m,s)=\mathbb{E}\left[\ell\left(Y,
  M_m(Q,R(Z_s,Q))\right)\right].
\end{equation}
The external-state gain relative to the foundation-only system is
\begin{equation}
  \gain(m,s)=L(m,0)-L(m,s).
\end{equation}
For ordered model sizes $m_1<m_2<m_3$ and a co-scaling rule $s(m)$, the
primary empirical property is
\begin{equation}
  \gain(m_1,s(m_1)) < \gain(m_2,s(m_2))
  < \gain(m_3,s(m_3)).
  \label{eq:scale-complementarity}
\end{equation}
We require $s(m)=o(m^2)$ and report both absolute state bytes and the ratio of
state bytes to model bytes. Equation~\eqref{eq:scale-complementarity} is a
property of the co-scaled system on a task distribution, not a universal model
law.

\begin{proposition}[No distribution-free strict scaling]
\label{prop:no-universal}
No fixed writer, reader, and co-scaling rule can guarantee strict positive
adjacent gain increments for every distribution over $(X,Q,Y)$.
\end{proposition}

\begin{proof}
Choose a distribution for which $Y$ is conditionally independent of $X$ given
$Q$, or for which the foundation-only predictor is already Bayes optimal. Then
every external state has zero Bayes gain. Alternatively, choose a state that
contains the exact answer at the smallest scale; bounded accuracy then prevents
strictly increasing absolute gain. Thus strict distribution-free scaling is
impossible.
\end{proof}

For log loss, define the Bayes risk of state $Z_s$ as
\begin{equation}
 L^*(s)=\inf_{p}\,\mathbb{E}[-\log p(Y\mid Q,Z_s)]
       =H(Y\mid Q,Z_s).
\end{equation}

\begin{proposition}[Information value of a refinement]
\label{prop:information-value}
If $Z_{s'}$ refines $Z_s$, the Bayes-risk improvement due to the refinement is
\begin{equation}
  L^*(s)-L^*(s') = I(Y;Z_{s'}\mid Q,Z_s).
\end{equation}
It is strict if and only if the refinement contains target-relevant conditional
information.
\end{proposition}

\begin{proof}
Because $Z_s$ is a function of $Z_{s'}$, the entropy chain rule gives
$H(Y\mid Q,Z_s)-H(Y\mid Q,Z_{s'})=I(Y;Z_{s'}\mid Q,Z_s)$.
\end{proof}

This identity supplies a constructive rule: each state expansion should add
nonredundant compositional evidence, while model scale determines how much of
that evidence a finite reader converts into reduced predictive loss.

\begin{proposition}[Strict complementarity on fresh episodic information]
\label{prop:oracle-complementarity}
Suppose the episode target is created after foundation training and is
conditionally independent of every foundation given $Q$. Let
$Z_1=\pi_1(Z_2)$ and $Z_2=\pi_2(Z_3)$ be nested state signals. If both the
no-state and state-conditioned predictors output their Bayes-optimal
conditionals and
\begin{equation}
 I(Y;Z_{i+1}\mid Q,Z_i)>0,\qquad i\in\{1,2\},
 \label{eq:nonredundant-refinement}
\end{equation}
then the Bayes log-loss gains satisfy
$\gain(m_1,s_1)<\gain(m_2,s_2)<\gain(m_3,s_3)$.
\end{proposition}

\begin{proof}
Fresh-information independence makes the no-state Bayes loss $H(Y\mid Q)$ at
every scale. Proper posterior decoding gives state loss $H(Y\mid Q,Z_i)$, so
the gain is $I(Y;Z_i\mid Q)$. By the chain rule and nesting,
\begin{equation}
 \gain(m_{i+1},s_{i+1})-\gain(m_i,s_i)
 =I(Y;Z_{i+1}\mid Q,Z_i)>0.
\end{equation}
\end{proof}

The indices $m_i$ in Proposition~\ref{prop:oracle-complementarity} carry no
force: with oracle decoders the gain depends only on the nested state. The
reader-dependent content of scale complementarity is isolated in
Proposition~\ref{prop:finite-reader}, whose interaction term is what the
factorial estimates.

The preceding statement concerns an oracle decoder. To connect it to frozen
finite readers, define the nonnegative excess risk
\begin{equation}
 \epsilon(m,s)=L(m,s)-H(Y\mid Q,Z_s).
\end{equation}
The observed gain admits the exact decomposition
\begin{equation}
 \gain(m,s)=I(Y;Z_s\mid Q)+\epsilon(m,0)-\epsilon(m,s).
 \label{eq:finite-reader-decomposition}
\end{equation}
It separates information present in the state from the reader's ability to
convert that information into predictive performance.

\begin{proposition}[Finite-reader interaction]
\label{prop:finite-reader}
For two frozen readers $m_a<m_b$ evaluated on the same nested refinement
$Z_s\preceq Z_{s'}$, let
\begin{equation}
 D_m(s,s')=\gain(m,s')-\gain(m,s).
\end{equation}
Then the model-by-state interaction is
\begin{align}
 D_{m_b}(s,s')-D_{m_a}(s,s')={}&
 [\epsilon(m_b,s)-\epsilon(m_b,s')] \nonumber\\
 &-[\epsilon(m_a,s)-\epsilon(m_a,s')].
 \label{eq:interaction-decomposition}
\end{align}
Consequently, a positive interaction identifies that the larger reader
realizes more of the same state refinement as excess-risk reduction.
\end{proposition}

\begin{proof}
Apply Eq.~\eqref{eq:finite-reader-decomposition} at $s$ and $s'$ for each
reader. The shared conditional-information increment
$I(Y;Z_{s'}\mid Q,Z_s)$ cancels between readers, leaving
Eq.~\eqref{eq:interaction-decomposition}.
\end{proof}

This identity motivates the full factorial rather than a diagonal-only
scaling curve. Diagonal contrasts establish the co-scaled system property;
the crossed interactions determine whether model capacity changes the value
extracted from an identical state expansion.

\begin{proposition}[Vanishing relative footprint]
\label{prop:footprint}
If state bytes obey $B(d)=O(d^\alpha)$ for $0<\alpha\leq1$, while a dense
$L$-layer foundation has $N=\Omega(Ld^2)$ parameters with $L=\Omega(1)$, then
$B(d)/N\to0$ as width $d\to\infty$.
\end{proposition}

\begin{proof}
Direct substitution yields
$B(d)/N=O(1/(Ld^{2-\alpha}))$; since $2-\alpha\geq1$ and $L$ is bounded below,
the ratio converges to zero.
\end{proof}

Thus the construction permits increasing absolute information value while the
relative state footprint vanishes. The finite-reader interaction experiment
tests the key empirical departure from the oracle premise: whether larger
frozen language models realize more of each identical nested refinement.

\section{ASCENT}

\begin{figure*}[t]
  \centering
  \includegraphics[width=\textwidth]{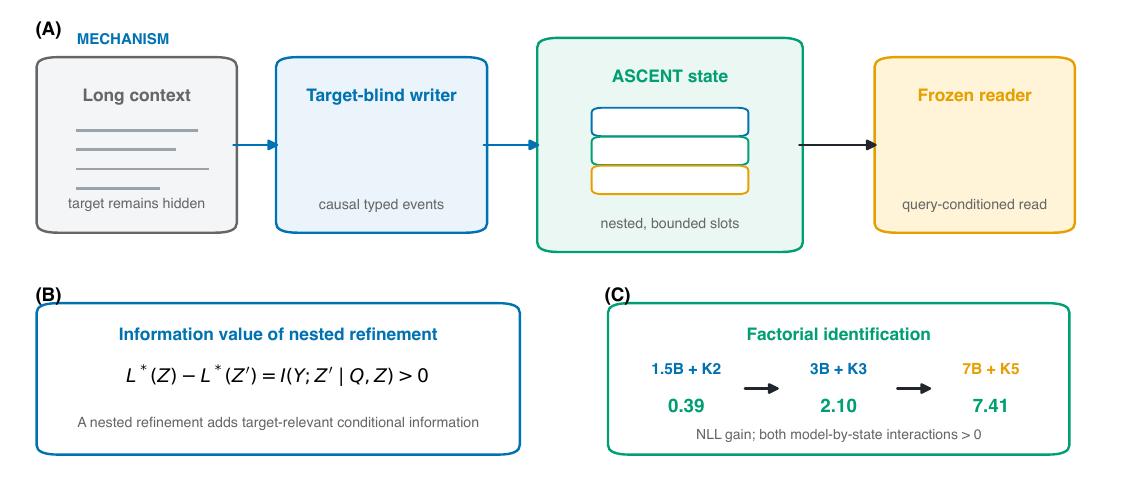}
  \caption{Theory, mechanism, and identification meet at the external-state
  gain. (A) A target-blind causal writer converts long evidence into nested,
  bounded state for query-conditioned use by a frozen reader. (B)
  Proposition~\ref{prop:information-value} equates the value of a strict
  refinement with its target-relevant conditional information. (C) The
  complete $3\times3$ factorial identifies the finite-reader consequence in
  Proposition~\ref{prop:finite-reader}: co-scaled gain rises from .39 to 2.10
  to 7.41 nats and both model-by-state interactions are positive.}
  \label{fig:ascent-method}
\end{figure*}

Figure~\ref{fig:ascent-method} summarizes the interface and the empirical
co-scaling target. The writer and foundation model remain frozen throughout;
the deterministic state budget alone follows the co-scaling rule. The design
therefore attributes the gain curve to reader capacity and state refinement,
with training held constant across endpoints.

\subsection{Execution Contract}

ASCENT is the deterministic composition
\begin{equation}
  Z_s=W_s(X),\qquad
  E_{s,Q}=R_s(Q,Z_s),\qquad
  \widehat{Y}_{m,s}=D_{M_m}\!\left(P(Q,E_{s,Q})\right),
  \label{eq:ascent-contract}
\end{equation}
where $W_s$ is the causal state writer, $R_s$ is a bounded readout, $P$ is a
fixed serializer, and $D_{M_m}$ is greedy decoding by the frozen reader. The
foundation arm is exactly $s=0$, for which $E_{0,Q}=\varnothing$. For
$s<s'$, a deterministic projection $\pi_{s'\rightarrow s}$ satisfies
$Z_s=\pi_{s'\rightarrow s}(Z_{s'})$, so every larger budget is a strict
refinement rather than a different representation.

Each endpoint executes the same five operations: (i) write state from $X$
before $Q$ and $Y$ are revealed; (ii) validate registered types, ordering,
hashes, alignment, and budget constraints; (iii) select at most $s$
query-relevant entries; (iv) serialize those entries with the fixed canonical
template; and (v) decode with the unchanged model and scorer. Validation is
fail-closed: a missing, malformed, or mismatched state aborts the invocation.
Equation~\eqref{eq:ascent-contract} isolates the two experimental
coordinates: $m$ changes only the frozen decoder, and $s$ changes only a
nested external-state refinement. It is the executable counterpart of
Propositions~\ref{prop:information-value} and~\ref{prop:finite-reader}.

\subsection{Target-Blind State Construction}

The writer processes the document before observing the final target. It
has three parts, which Table~\ref{tab:writer-structure} lists for every
evaluated document family. A \emph{recognizer} scans the input that
precedes the question and emits typed records; each record carries its
fields, its character position, and the verbatim input span from which it
was read. A \emph{state update} folds the records, in document order, into
bounded state built from three primitives: a last-write map, an append-only
log, and a counter. A \emph{question map} identifies the question form and
names the state keys from which the readout may select. The record types,
their composition into state, and the question forms constitute a family's
schema; across the evaluated families the schemas record typed events,
assignment edges, occurrence-indexed key--value evidence, and count
summaries. Everything outside the schema is common to all families: the
write boundary at the question, position-and-span provenance, registered
slot bounds, nested budgets fixed before evaluation, and fail-closed
validation.

Two of the four families are not bAbI-style. RULER multi-query retrieval
embeds key--value needles in essay text, and RULER common-word aggregation
presents a numbered list of words; for each, the recognizer is a single
regular expression and the state a single primitive, a key--value map or an
exact counter. Adapting ASCENT to a new document family therefore amounts
to supplying its schema. The rest of the pipeline sees a recognizer only
through its output, so any extractor, rule-based or learned, that emits
position-anchored verbatim records before the question reuses the same
state primitives, budget, and validation unchanged.

\begin{table}[t]
\caption{Writer structure for each evaluated document family. The
recognizer, the state composition, and the question map form the family's
schema; the write boundary, provenance, slot bounds, nested budgets, and
fail-closed validation are shared. Bounds and budgets are the registered
configuration values, listed in order of reader size.}
\label{tab:writer-structure}
\centering
\scriptsize
\setlength{\tabcolsep}{3pt}
\begin{tabular}{@{}>{\raggedright\arraybackslash}p{.13\textwidth}>{\raggedright\arraybackslash}p{.13\textwidth}>{\raggedright\arraybackslash}p{.21\textwidth}>{\raggedright\arraybackslash}p{.19\textwidth}>{\raggedright\arraybackslash}p{.12\textwidth}>{\raggedright\arraybackslash}p{.13\textwidth}@{}}
\toprule
Family & Document & Recognizer $\to$ record & State (bound) & Question $\to$ keys & Budget $s(m)$ \\
\midrule
BABILong QA1--QA3 & Book text with interleaved event sentences & 4 patterns over closed person, place, and object sets $\to$ move, pick-up, drop, transfer (person; place, object, recipient) & Last-write maps person$\to$place and object$\to$holder, joined to object$\to$place; per-object place log (256) & 3 forms $\to$ person; object; object and place & 1/2/3 fact slots; 4 at 7B--8B \\
\addlinespace
BABILong QA7--QA8 & As above & Same records & Per-person inventory log; count of held objects & 2 forms $\to$ person & 1/2/8 fact slots \\
\addlinespace
RULER multi-query retrieval & Essay text with key--value needles & 1 pattern $\to$ (key, value) & Last-write key$\to$value map (256) & Keys named in the question & 40/69/91 replay tokens (Qwen2.5); 37/55/97 (SmolLM2) \\
\addlinespace
RULER common-word aggregation & Numbered list of words & 1 pattern $\to$ list item & Exact counter (512 or 1,024; overflow aborts) & Frequency rank & 3/7/10 words (Qwen2.5); 3/5/10 (SmolLM2) \\
\bottomrule
\end{tabular}
\end{table}

The implementation contains no instance answer. Parser checks precede model
decoding, and state entries preserve document order under deterministic
tie-breaking. Each released BABILong result row records the fields the
writer read, that the reference answer was inaccessible until the prompt had
been rendered, and that every retained fact is a verbatim span of the input
at its recorded position (Section~\ref{sec:repro}).

\subsection{Co-Scaled Budget}

The budget $s(m)$ is fixed before evaluation. It may grow in the number of
events, slots, or replay tokens, but remains small relative to model storage.
Fixed-state arms separate larger-reader effects from richer-state effects; the
full factorial estimates their interaction.

\subsection{Readout and Representation}

The reader supplies only query-relevant state. For event-state QA, a canonical
deictic-free form separates semantic entities, relations, and temporal order.
Matched representation experiments identify canonical serialization as the
effective interface: at 3B it raises official 16K accuracy by 15.38 points over
raw event wording while leaving every Foundation prediction unchanged. The
canonical interface was selected on an 8K panel and fixed before official 16K
evaluation.

\subsection{Implementation Details}

For the principal BABILong experiment, the writer extracts occurrence-indexed
entities, relations, and temporal events from the context before the question
is presented to the model. The budgets are one, two, and three
query-relevant fact slots for the 0.5B, 1.5B, and 3B Qwen2.5 readers,
respectively. The readout uses a fixed canonical serialization and at most eight
new decoding tokens. Models are loaded from pinned revisions, are never
fine-tuned, and decode greedily. Foundation and \ascent{} predictions share the
same tokenizer, answer normalizer, and scoring implementation.

\section{Experimental Protocol}

\subsection{Models and Benchmarks}

We use frozen Qwen2.5-Instruct \cite{qwen2024qwen25} readers at 0.5B, 1.5B, 3B,
and 7B parameters and SmolLM2 \cite{allal2025smollm2} readers at 135M, 360M,
and 1.7B parameters. A separately frozen breadth extension contains Mistral-7B-Instruct-v0.3
\cite{jiang2023mistral}, Falcon3-7B-Instruct \cite{tii2024falcon3},
Granite-3.3-8B-Instruct \cite{ibm2025granite33}, and Qwen3-8B in nonthinking
mode \cite{yang2025qwen3}. Official RULER tasks cover multi-query
retrieval and common-word aggregation. Official BABILong tasks cover retrieval,
composition, and aggregation at 4K, 8K, and 16K contexts. All decoding is greedy
and uses the same normalization path within a comparison.

\subsection{Experimental Design and Inference}

Configurations, panel hashes, model revisions, decision rules, and analysis
code are fixed before scores are inspected. The main Qwen2.5 16K study
contains ten panels, each with 40 QA2 and 40 QA3 examples.
The two directional adjacent gain contrasts form one family and are tested with
panel-clustered $t$ statistics and Holm family-wise error control at 0.05. All
panels and tasks are retained.

The panel, rather than the individual question, is the unit of inference. For
each endpoint, we first average QA2 and QA3 accuracy within each of ten
independently selected panels and then compute the paired
\ascent{}--Foundation gain. Student-$t$ intervals use the ten panel estimates
($9$ degrees of freedom). The two one-sided adjacent gain-increment tests were
specified before decoding and corrected jointly by Holm's step-down procedure.
Endpoint intervals and two-sided adjacent intervals characterize effect size;
the prespecified directional family supplies the primary decision rule.

A separate mechanism study crosses Qwen2.5 1.5B, 3B, and 7B with nested
state sizes $K\in\{2,3,5\}$. Interface selection uses one independent panel;
the factorial analysis uses nine disjoint panels, each containing 256 new
examples from an independent categorical observation channel. Two latent
values in $\{0,\ldots,7\}$ must be
composed by ordinary addition. To avoid parse and arbitrary-token-frequency
confounds, we score the exact joint likelihood of the 15 equal-length two-token
continuations \texttt{00},\ldots,\texttt{14} after one fixed assistant prefix.
The fifteen continuations tokenize to equal length at every row and endpoint.
The gain is the reduction in candidate-normalized target NLL versus Foundation.
In addition to three co-scaled gains and two adjacent increments, the
interactions compare the value of $K_2\!\to K_3$ at 3B versus 1.5B and the value
of $K_3\!\to K_5$ at 7B versus 3B. All eight two-sided $t_8$ intervals must have
positive lower bounds; a Bonferroni family supplies simultaneous 95\%
intervals over all eight contrasts.

The semantic validation set is selected score-blind from the public
BABILong training-generation split. We exclude any row whose task, normalized
question, and target-blind parsed-event sequence matches any row in the full
official 8K test pool. The resulting 800 rows have zero such semantic
fingerprint overlap. It supplies a distribution-matched replication alongside
the official held-out evaluation.

\subsection{Baselines and Controls}

The baseline suite includes the foundation-only reader; fixed versus co-scaled
state; exact and byte-matched FIFO memory; lexical retrieval; BGE-M3 dense plus
BM25 reciprocal-rank fusion and reranking; same-FLOP static replay; irrelevant
and corrupted state; certified-only, latent-only, and fused paths; and released
task-trained Q-RAG QA2/QA3 checkpoints. Q-RAG is evaluated on the same 800
examples and with the same frozen Qwen readers, preserving its task-specific
training advantage.

\subsection{Theory--Experiment Correspondence}

Table~\ref{tab:theory-evidence} links each principal claim to a distinct
observable quantity. Exact-posterior transitions establish
Proposition~\ref{prop:information-value}. Diagonal contrasts test the co-scaled
system property, while crossed interactions identify
Proposition~\ref{prop:finite-reader} at matched state refinements. Official
generation establishes the task-level consequence. Systems measurements
instantiate Proposition~\ref{prop:footprint}.

Nested state differences estimate the information supplied by refinement;
adjacent diagonal differences test Eq.~\eqref{eq:scale-complementarity};
crossed second differences isolate the reader-dependent extraction term in
Eq.~\eqref{eq:interaction-decomposition}; and normalized state bytes test the
footprint statement. Each theoretical operator has its own observable.

\begin{table}[H]
\caption{Theory--experiment correspondence. Each proposition is paired with
its identifying estimand and empirical confirmation. Mechanism intervals are
Bonferroni-simultaneous over the eight-estimand family; official accuracy uses
the prespecified Holm-corrected directional family.}
\label{tab:theory-evidence}
\centering
\scriptsize
\setlength{\tabcolsep}{4pt}
\begin{tabular}{>{\raggedright\arraybackslash}p{.15\textwidth}>{\raggedright\arraybackslash}p{.25\textwidth}>{\raggedright\arraybackslash}p{.33\textwidth}>{\raggedright\arraybackslash}p{.12\textwidth}}
\toprule
Theoretical claim & Empirical quantity & Quantitative result & Conclusion \\
\midrule
Refinement value (Prop.~\ref{prop:information-value}) & Exact-posterior loss under each nested $K$ transition & 13,824 of 13,824 endpoint--transition row checks: each larger state extends the smaller one and adds an observation (true by construction); panel-mean exact-posterior NLL decreases on every transition in every panel. & Design check\footnotemark[1] \\
Co-scaled complementarity (Eq.~\ref{eq:scale-complementarity}) & Diagonal gain and two adjacent increments & .388$\to$2.099$\to$7.406 nats; increments 1.711 [1.540,1.882] and 5.307 [5.053,5.561]. & Established \\
Finite-reader interaction (Prop.~\ref{prop:finite-reader}) & Two crossed interaction contrasts & .781 [.539,1.023] and .968 [.671,1.264]; simultaneous 95\% CIs (Bonferroni). & Identified \\
Task-level consequence & Official 16K generation accuracy gain & 10.50$\to$78.88$\to$82.63 points; both directional contrasts remain significant after Holm correction. & Established \\
Relative footprint (Prop.~\ref{prop:footprint}) & External-state bytes and state/model ratio & 589.0$\to$624.6$\to$640.6 bytes while the ratio falls $2.189\!\times\!10^{-6}\to8.631\!\times\!10^{-7}\to1.872\!\times\!10^{-7}$. & Established \\
\bottomrule
\end{tabular}
\end{table}
\footnotetext[1]{Under the nested construction each additional slot extends
the retained observation prefix of the same latent registers, so the row
checks (prefix nesting and non-redundancy across all 13,824
endpoint--transition rows) hold by construction. They verify the
implementation and are not independent empirical evidence. Per row, the exact
posterior improves on 10,791 of the 13,824 transitions under the noisy
channel (\ref{app:power}).}

\section{Results}
\label{sec:results}

\subsection{Qwen2.5 Scaling on Official BABILong}

\begin{figure*}[t]
  \centering
  \includegraphics[width=\textwidth]{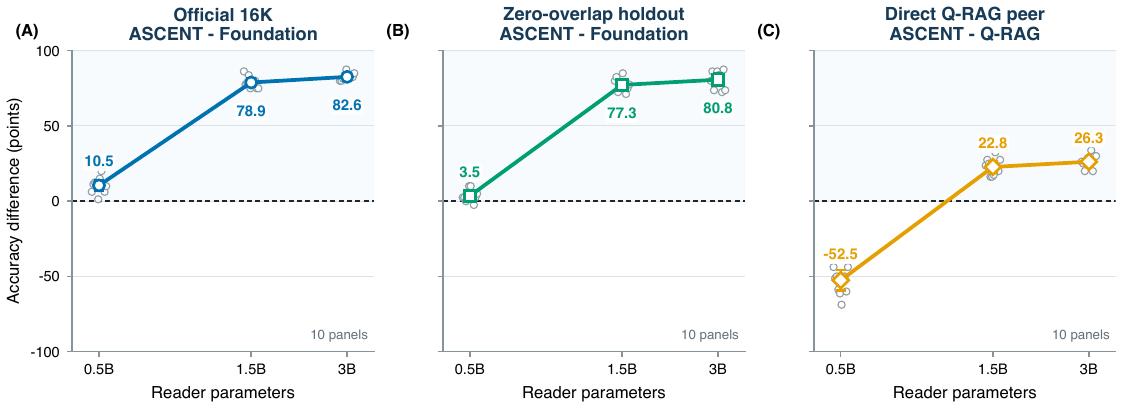}
  \caption{Scale-complementary gains across three ordered Qwen2.5 readers.
  Gray points are the ten task-panel clusters; colored symbols are means and
  vertical bars are 95\% Student-$t$ intervals. (A) Official 16K BABILong.
  (B) Score-blind validation with zero semantic-fingerprint overlap with the
  pinned official pool. (C) Direct comparison with released task-trained
  Q-RAG on identical rows. The official and zero-overlap curves reproduce the
  same strict adjacent ordering; the peer margin crosses zero at 1.5B and
  widens at 3B.}
  \label{fig:qwen-primary-evidence}
\end{figure*}

Figure~\ref{fig:qwen-primary-evidence} shows every panel-level observation.
Both evaluation families produce the same sharply increasing curve. The
direct-peer margin becomes positive at 1.5B and increases further at 3B.

Table~\ref{tab:qwen-confirmations} reports the two prespecified Qwen2.5
curves. The official 16K curve increases from 10.50 to 78.88 to 82.63
percentage points. The adjacent increments are 68.38 points
($p_{\mathrm{Holm}}=3.43\times10^{-11}$) and 3.75 points
($p_{\mathrm{Holm}}=0.0255$). Both adjacent directional contrasts remain
significant under family-wise correction; the second is significant but
marginal (\ref{app:power}).

The semantic-holdout curve increases from 3.50 to 77.25 to 80.75 points. Its
adjacent increments are 73.75 points
($p_{\mathrm{Holm}}=1.65\times10^{-13}$) and 3.50 points
($p_{\mathrm{Holm}}=0.02897$), reproducing both positive adjacent increments.

\begin{table}[t]
\caption{Qwen2.5 task-balanced QA2+QA3 accuracy. Intervals are panel-clustered
95\% intervals for \ascent{}--Foundation gain.}
\label{tab:qwen-confirmations}
\centering
\footnotesize
\resizebox{\columnwidth}{!}{%
\begin{tabular}{llccc}
\toprule
Split & Reader & Foundation & \ascent{} & Gain [95\% CI] \\
\midrule
\multirow{3}{*}{Official 16K} & 0.5B & .0563 & .1613 & .1050 [.0680,.1420] \\
 & 1.5B & .1888 & .9775 & .7888 [.7616,.8159] \\
 & 3B & .1613 & .9875 & .8263 [.8097,.8428] \\
\midrule
\multirow{3}{*}{Semantic holdout} & 0.5B & .1263 & .1613 & .0350 [.0065,.0635] \\
 & 1.5B & .2025 & .9750 & .7725 [.7419,.8031] \\
 & 3B & .1763 & .9838 & .8075 [.7668,.8482] \\
\bottomrule
\end{tabular}
}
\end{table}

At 3B on the official 16K panels, the canonical representation reaches .9875
accuracy versus .8338 for the frozen raw representation, a 15.38-point
difference with 95\% CI [13.69,17.06]. Foundation predictions are identical
row-by-row, isolating the state representation.

\subsection{Direct Full-Factorial Scale Interaction}

\begin{figure*}[t]
  \centering
  \includegraphics[width=\textwidth]{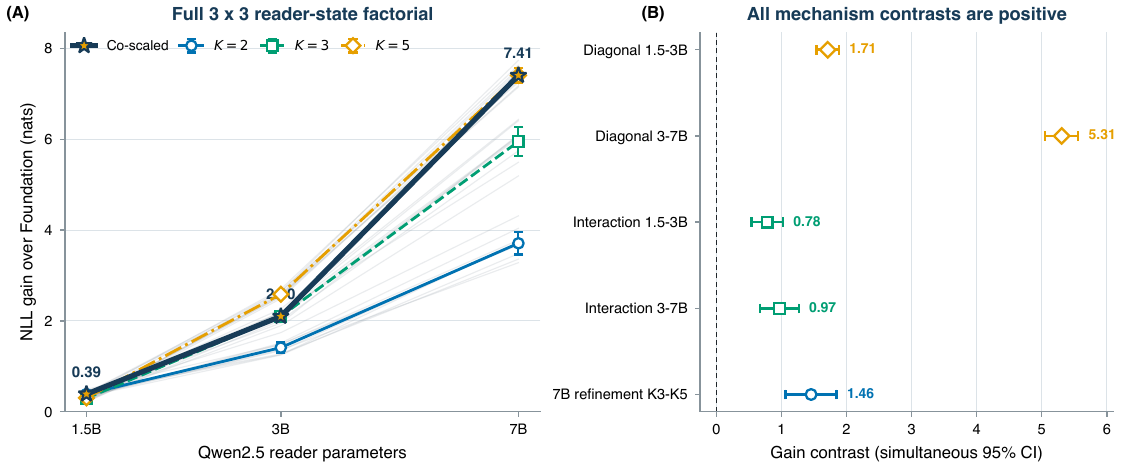}
  \caption{Direct identification of scale complementarity. (A) The light
  trajectories expose all nine independent panels; colored summaries give the
  mean and 95\% interval in every reader--state cell, and the navy starred line
  is the co-scaled diagonal. (B) Bonferroni-simultaneous 95\% intervals for the
  two diagonal increments, two model-by-state interactions, and largest-reader
  refinement. Their positive lower bounds jointly establish the diagonal
  property and the finite-reader mechanism in
  Proposition~\ref{prop:finite-reader}.}
  \label{fig:factorial}
\end{figure*}

Figure~\ref{fig:factorial} reports the complete factorial mechanism study.
The co-scaled NLL gains increase from .388 to 2.099 to 7.406 nats
(Fig.~\ref{fig:factorial}). Their adjacent increments are 1.711
[1.540,1.882] and 5.307 [5.053,5.561]. Both model-by-state interactions are
positive: .781 [.539,1.023] and .968 [.671,1.264]. These Bonferroni-simultaneous
intervals retain positive lower bounds over the full eight-estimand family.
Exact nesting, target isolation, Foundation
probability identity across $K$, and all 13,824 endpoint--transition row checks
hold across the factorial. The row checks hold by construction of the nested
writer (Table~\ref{tab:theory-evidence}, note); the per-row exact-posterior
tallies are in \ref{app:power}. Reruns of factorial panels on other
instances are reported in Section~\ref{sec:repro}.

\subsection{Public 7B--8B Breadth}

\begin{figure*}[t]
  \centering
  \includegraphics[width=\textwidth]{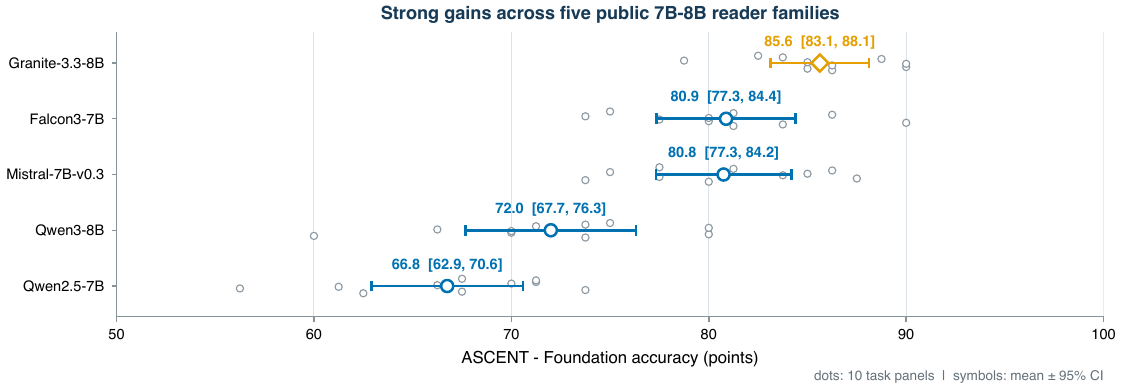}
  \caption{Official 16K BABILong breadth across five public 7B--8B readers.
  Gray points are the ten task-panel gains; symbols are means and bars are
  panel-clustered 95\% intervals. All five architectures realize large
  positive gains, extending the task-level consequence beyond the Qwen2.5
  scaling family.}
  \label{fig:breadth}
\end{figure*}

Every planned public 7B--8B reader has complete ten-panel official 16K
BABILong QA2/QA3 results (Fig.~\ref{fig:breadth}). All five gain intervals
exclude zero: Qwen2.5-7B gains 66.75 points, Qwen3-8B 72.00, Mistral-7B 80.75,
Falcon3-7B 80.88, and Granite-3.3-8B 85.63. The gain persists across five
architectures in the 7B--8B range.

\subsection{Cross-Family and Cross-Task Evidence}

The cross-task evidence matrix extends the primary result across model
families, task structures, and context lengths. For SmolLM2 on official RULER
common-word aggregation, gains increase from
.2628 to .4378 to .9561, with both adjacent intervals strictly positive.
Qwen2.5 at 16K also has increasing means, .2417 to .6144 to .7100, with a
positive log-scale slope. On official RULER multi-query retrieval,
both Qwen2.5 (.0438 to .1208 to .3188) and SmolLM2 (.1042 to .2167 to .5229)
have positive intervals for both adjacent increments. Official BABILong 4K
and 8K studies cover QA1--QA3, while QA7--QA8 provide
aggregation evidence.

\subsection{Direct Current-Peer Comparison}

Against released task-trained Q-RAG checkpoints on identical QA2/QA3 panels,
frozen \ascent{} leads by 22.75 points [.1885,.2666] at 1.5B and by 26.25
points [.2307,.2943] at 3B. The increasing positive margin at the two larger
shared scales demonstrates that scale-complementary state can surpass
task-trained retrieval without updating the foundation reader.

\subsection{Controls and Systems Evidence}

ASCENT outperforms byte-matched FIFO, lexical retrieval, and BGE-M3 hybrid
retrieval on matched SmolLM2 panels. Irrelevant and corrupted-memory controls
isolate the benefit of structured evidence from prompt length. Across 18 clean processes, median
isolated decode-time ratios are .4130, .4247, and .2791 for the three ordered
readers. Fixed write-plus-read state occupies 589.0, 624.6, and 640.6 bytes,
while the state-to-model byte ratio decreases with scale. Supported-operator
profiles show 83.7$\times$, 79.4$\times$, and 78.2$\times$ fewer FLOPs than the
foundation generate path.

\section{Limitations}
\label{sec:limitations}

\paragraph{Scope of the claim} Scale complementarity is demonstrated here for
Qwen2.5 readers on official BABILong and RULER tasks, under accuracy and under
candidate-normalized negative log-likelihood, with a structured-fact state
schema. It is not shown to be a general property of frozen long-context
readers with external memory. Within this scope three results are
unfavorable and are reported as such: the accuracy gain decreases from 3B to
7B; the upper adjacent interval for Qwen2.5 common-word aggregation includes
zero; and the SmolLM2 same-refinement interaction intervals include zero.

\paragraph{Schema scope} The writer schema used here
(Table~\ref{tab:writer-structure}) records typed events,
assignment edges, occurrence-indexed key--value evidence, and count summaries.
These are the structures present in the evaluated structured-fact-retrieval
benchmarks. The present evidence therefore establishes scale complementarity
for external state constructed against such structures, and does not
establish it as a general property of frozen readers with arbitrary external
memory. \ref{app:schema-blind} reports the identical pipeline with
three schema-free writers (a sentence-recency window, a lexical retrieval
chain, and BGE-M3 hybrid retrieval with reranking) in place of the
structured-fact writer. None satisfies the registered co-scaling rule: at
the same one-to-three-slot budget the generic state is worse than no state
at every scale (gains of $-5$ to $-18$ points), and the structured-fact
writer's advantage over each generic writer is 15 points at 0.5B and 89--99
points at 1.5B and 3B. The pattern reported here is therefore a property of
external state constructed against this schema. Extending the writer to
non-templated natural text is the principal open direction.

\paragraph{Unfavorable and marginal results} The claim is specific to
scale and estimand. Official BABILong accuracy
establishes increasing gains from 0.5B to 3B; the second increment
(1.5B$\to$3B, 3.75 points) is significant but marginal after Holm correction
($p_{\mathrm{Holm}}=0.0255$), and \ref{app:power} shows that the
ten-panel design detects an increment of this size with power of about .52,
against a minimum detectable increment of about 5.3--5.8 points at 80\%
power. At 7B,
Foundation accuracy rises to .3325 and four-slot \ascent{} reaches its ceiling;
the gain is .6675, yielding a 3B-to-7B increment of $-.1588$
[$-.2055,-.1120$]. The 7B conclusion therefore rests on the non-saturated NLL
factorial, where both model-by-state interactions have positive simultaneous
lower bounds. Likewise, Qwen2.5 common-word aggregation establishes a positive
scaling slope, while its upper adjacent interval is .0956
[$-.0043$,.1954]. The task-trained Q-RAG comparison favors ASCENT at 1.5B and
3B, but not at 0.5B. These cases match Proposition~\ref{prop:no-universal}:
complementarity requires both informative refinement and reader headroom. The
SmolLM2 factorials reproduce increasing co-scaled gains, while their
same-refinement interaction intervals overlap zero; the Qwen NLL factorial
therefore supplies the direct reader-by-state interaction result.

The architecture study covers autoregressive Transformers rather than
recurrent or linear readers. The distribution-matched semantic set originates
from a training-generation split and is consequently secondary to the official
held-out results. The direct interaction study uses a controlled compositional
task and teacher-forced likelihood, so natural-text generation remains an
important extension. Finally, the systems result establishes bounded state,
decreasing relative storage, isolated decode gains, and 78.2--83.7$\times$
supported-operator FLOP reductions; its profiler-enclosed end-to-end
wall-time ratios of .9008, .8651, and .7262 are reported as measured.

\section{Reproducibility and Ethics}
\label{sec:repro}

\paragraph{Public record} The code, frozen configurations, per-row result
records, analysis files, and audit reports are released at
\url{https://github.com/XingjianZhang-dev/ascent} and archived at Zenodo
(version DOI \path{10.5281/zenodo.22865847}). One CPU command
(\texttt{make reproduce}) re-verifies the SHA-256 manifests, recomputes every
analysis file from the per-row records, re-renders every table in this paper
byte for byte, and prints every reported number; \texttt{VERIFICATION.md}
maps each value to its records. The freeze commit precedes the first score
for all 25 frozen configurations with retained scores
(\path{PREREGISTRATION_TIMELINE.md}).

\paragraph{Reproduction on other hardware} Every formal run executed on
NVIDIA RTX PRO 6000 Blackwell instances. Reruns on matched hardware
reproduced three audit cells identically, at the level of prediction and,
where floats are stored, IEEE-754 binary64 identity: the factorial
development panel (256 records, node 1 versus node 2), official 16K panel 1
for Qwen2.5-7B (80 rows, node 1 versus node 2), and all ten official 16K
panels for Qwen3-8B (800 rows, third instance). The three instances have
distinct GPU serial numbers, and the first two have distinct host driver
versions. The Qwen3-8B rerun is a post-hoc code audit and enters no reported
number; it ran from a one-commit repository snapshot whose working tree also
held uncommitted manuscript-packaging edits, none of which the runner
imports, and the diff is retained in the repository. Every other formal cell
was run once.

A rerun on a separately provisioned instance with a different GPU
architecture (NVIDIA A100, Ampere) and identical software pins reproduced
the reported conclusions; stored values were not bit-identical. Across all
sixty official-16K and semantic-holdout cells
(4,800 rows, both arms), 124 of 9,600 scored outputs changed (103 Foundation,
21 \ascent{}); twenty cells were unchanged, and single 80-row panels moved by
up to 5.00 accuracy points (four rows). The ten-panel means moved by at most
1.00 point (official 16K: 10.50/78.88/82.63 $\to$ 9.75/78.88/83.25 points;
semantic holdout: 3.50/77.25/80.75 $\to$ 4.50/77.00/80.63), inside every
reported interval half-width (1.66--3.70 points), and every registered
directional contrast still passes (second official increment 4.38 points,
$p_{\mathrm{Holm}}=0.0028$; semantic holdout 3.63 points,
$p_{\mathrm{Holm}}=0.022$). On factorial panels 2 and 3, one to two of 256
argmax candidates per arm flipped and the panel NLL gains moved by
approximately 0.01 nats against a reported half-width of 0.16 nats. Stored
floating-point values differ in their low-order bits, as expected across
kernel implementations. Same-instance controls were bit-identical, so the
differences follow from the architecture change. The numbers reported in
this paper are the Blackwell values.

\paragraph{Determinism} Decoding is greedy in bfloat16 with pinned
weights, revisions, panels, seeds, and batch sizes;
\path{torch.use_deterministic_algorithms} and a TF32 policy were not set
(\path{reproduction/DETERMINISM_POLICY.md}). The flag requests algorithms
that return the same output for the same input when run on the same
software and hardware \cite{pytorch2025deterministic}. The pipeline
already has that property without it: the three matched-hardware audit
cells, two complete executions of the pre-registered A100 audit cells, and
six A100 cells run with and without the target-blindness instrumentation are
each identical at the prediction level and, where floats are stored, at
binary64. Strict
same-hardware determinism was thus obtained with the default kernels, at
the throughput of the runs as executed; the flag was not needed, and a
flag-enabled run was not timed. The flag does not extend across GPU
architectures. cuBLAS guarantees bit-wise identical results only on GPUs
with the same architecture and the same number of streaming
multiprocessors \cite{nvidia2025cublas}, and PyTorch does not guarantee
reproducibility across platforms \cite{pytorch2025reproducibility}; the
Blackwell and A100 instances execute architecture-specific
matrix-multiplication and attention kernels whose floating-point reductions
need not round identically. The 124 changed outputs arise from this
difference, and enabling the flag would not have removed them. Agreement
across architectures is accordingly assessed where the conclusions are
stated, in panel means and registered contrasts, and it holds there.

\paragraph{Target-blindness} The BABILong writer never receives the target;
it is called with the input and the question only. The writer, readout, and
prompt builder run behind a blinded row view that raises on any access to
the answer field until the prompt has been rendered, and every result row
records the fields read, the number of blocked accesses (zero), and a
provenance check that each retained fact is the verbatim input span at its
recorded position. All 4,800 rows of the official 16K and semantic-holdout
studies were run with this instrumentation; on the same instance the
instrumented and uninstrumented runners produced identical outputs.

\paragraph{Interruptions and licenses} The RULER common-word
confirmation had two pre-score interruptions, a missing legacy
configuration field and a node without the SmolLM2-1.7B weights; neither
produced a score, and both are logged in the release. The released per-row
records omit benchmark row bodies and model weights, which remain under
their original licenses; panel identifiers, hashes, pinned revisions, and
construction scripts are released instead.

\section{Conclusion}

On the evaluated tasks, external state becomes more valuable as a frozen
language-model reader scales.
Conditional information theory specifies the value of nested refinement, and
the factorial interaction shows that larger readers extract increasingly more
of that value. Official RULER and BABILong evaluations reproduce the
increasing-gain pattern across model families, tasks, and context lengths; five
public 7B--8B readers retain large absolute gains. Direct-peer comparisons,
systems measurements, representation controls, and cross-node reproduction
complete the theory-to-evidence chain. Scale complementarity is therefore
identifiable, achievable, and reproducible in the evaluated frozen
long-context language-model systems, within the scope stated in
Section~\ref{sec:limitations}.

\section*{CRediT authorship contribution statement}
  Xingjian Zhang: Conceptualization, Methodology, Software, Formal analysis,
  Investigation, Validation, Data curation, Visualization, Writing -- original
  draft, Writing -- review and editing, Project administration.

  \section*{Funding}
  This research did not receive any specific grant from funding agencies in the
  public, commercial, or not-for-profit sectors.

  \section*{Declaration of competing interest}
  The author declares that he has no known competing financial interests or
  personal relationships that could have appeared to influence the work reported
  in this paper.

\section*{Data availability}
Code and analysis artifacts are publicly available at
\url{https://github.com/XingjianZhang-dev/ascent} and archived at
\url{https://doi.org/10.5281/zenodo.22865847}. Panel identifiers, hashes,
and construction scripts are released for BABILong and RULER; the underlying
benchmark corpora are not redistributed and remain under their original
licenses (Apache-2.0 and BSD components; see \path{THIRD_PARTY_NOTICES.md} in the repository).
Third-party model weights are not redistributed and remain subject to their
original licenses.

\section*{Declaration of generative AI and AI-assisted technologies in the
manuscript preparation process}
During the preparation and revision of this work, the author used ChatGPT
and Codex and, during the revision, Claude Code as
assistive tools for implementing author-specified code modifications,
conducting routine checks, and improving the clarity and wording of the
manuscript and response to reviewers. No AI tool produced any reported
measurement. All research questions, methodological decisions, experimental
design, validation, interpretation of results, and final written content were
determined, reviewed, and approved by the author, who takes full
responsibility for the work. A per-file record is provided as
\texttt{AI\_ASSISTANCE.md} in the repository.

\appendix
\numberwithin{table}{section}
\renewcommand{\thetable}{\Alph{section}.\arabic{table}}

\section{Complete Factorial Cell Table}
\label{app:cells}

Table~\ref{tab:app-cells} lists the mean candidate-normalized NLL gain and its
95\% Student-$t_8$ interval in every reader$\times$state cell of the numeric
factorial (nine independent 256-row panels per cell; Fig.~\ref{fig:factorial}
plots the same cells). Starred cells form the co-scaled diagonal. Reading
along a row isolates the state-budget component at fixed reader; reading down
a column isolates the reader component at fixed state; the diagonal
increments and the two crossed interactions in Table~\ref{tab:app-mechanism-r1}
are second differences of these entries.

\begin{table}[H]
\caption{Mean NLL gain (nats) with 95\% $t_8$ intervals in all nine
reader$\times$state cells. $\star$ marks the co-scaled diagonal.}
\label{tab:app-cells}
\centering
\small
\begin{tabular}{lccc}
\toprule
Reader & $K=2$ & $K=3$ & $K=5$ \\
\midrule
1.5B & .388 [.340,.437]$\star$ & .295 [.247,.342] & .304 [.248,.359] \\
3B & 1.412 [1.288,1.535] & 2.099 [2.004,2.194]$\star$ & 2.587 [2.536,2.638] \\
7B & 3.709 [3.460,3.958] & 5.950 [5.637,6.264] & 7.406 [7.246,7.566]$\star$ \\
\bottomrule
\end{tabular}
\end{table}

\begin{table}[H]
\caption{Registered factorial estimands with Bonferroni-simultaneous 95\%
intervals over the eight-estimand family
($t_{1-.05/16,8}=3.677$).}
\label{tab:app-mechanism-r1}
\centering
\small
\begin{tabular}{lccc}
\toprule
Reader & State $K$ & Gain (nats) & Simultaneous 95\% CI \\
\midrule
1.5B & 2 & .388 & [.311,.465] \\
3B & 3 & 2.099 & [1.948,2.251] \\
7B & 5 & 7.406 & [7.151,7.661] \\
\midrule
Model$\times$state: 1.5B$\to$3B & -- & .781 & [.539,1.023] \\
Model$\times$state: 3B$\to$7B & -- & .968 & [.671,1.264] \\
\bottomrule
\end{tabular}
\end{table}

\section{Statistical Formulation, Power, and Consistency}
\label{app:power}

\paragraph{Estimands} Let $g_{m,k,p}$ denote the paired Foundation-minus-\ascent{}
loss difference for reader $m$, state size $k$, and panel $p$. The co-scaled
curve is $\bar g_{1.5,2},\bar g_{3,3},\bar g_{7,5}$; the diagonal increments
are its adjacent differences; the interactions are
$\delta_1=(\bar g_{3,3}-\bar g_{3,2})-(\bar g_{1.5,3}-\bar g_{1.5,2})$ and
$\delta_2=(\bar g_{7,5}-\bar g_{7,3})-(\bar g_{3,5}-\bar g_{3,3})$; the eighth
estimand is $\bar g_{7,5}-\bar g_{7,3}$. All quantities are computed within
panel and averaged over nine panels; ordinary intervals use $t_8$ and
simultaneous intervals use $t_{1-.05/16,8}$. The official Qwen2.5 BABILong
study registers two one-sided adjacent accuracy-gain contrasts over ten
panels with Holm step-down control at .05; the smallest Holm threshold is
therefore $.025$ one-sided. No row, panel, endpoint, or outlier is removed.

\paragraph{Retrospective power} Table~\ref{tab:app-power} recomputes, from the
panel-level evidence files released with the code, the between-panel SD, SE,
$t$ statistic and raw one-sided $p$ of every registered contrast, the minimum
detectable effect (MDE) at the registered per-test level and 80\% power
(exact non-central $t$), and the achieved power for the observed effect. Two
facts follow. The official-16K design is well powered for the first
increment ($t_9=37.4$) and underpowered for the second: with the observed
between-panel SD, the MDE at 80\% power is 5.25 points on the second
contrast's SE (5.75 on the first's), so the observed 3.75-point 1.5B$\to$3B
increment was detected at power $\approx.52$ and is described as
significant but marginal ($p_{\mathrm{Holm}}=.0255$). The very small
$p$-values elsewhere follow from large effects with small between-panel
variance (Table~\ref{tab:app-consistency}: a 68.38-point increment with
SE 1.83 points gives $t_9=37.4$). Retrospective power uses the observed SE
as if it were the true SD; it characterizes the sensitivity of the design,
not the probability that a result is true.

\begin{table}[H]
\caption{Registered contrasts: effect, SE, $t$, raw one-sided $p$, minimum
detectable effect at the registered per-test level and 80\% power, and
achieved power. Official-16K and semantic-holdout rows use $t_9$ and the Holm
step level $.025$; factorial rows use $t_8$ and the Bonferroni level
$.05/16$ one-sided.}
\label{tab:app-power}
\centering
\scriptsize
\setlength{\tabcolsep}{3pt}
\resizebox{\columnwidth}{!}{%
\begin{tabular}{llrrrrrr}
\toprule
Study & Contrast & Effect & SE & $t$ & $p_{\mathrm{raw}}$ & MDE$_{80}$ & Power \\
\midrule
Official 16K & 0.5B $\to$ 1.5B & 68.38 pts & 1.83 & 37.44 & $1.72\times10^{-11}$ & 5.75 pts & 1.00 \\
Official 16K & 1.5B $\to$ 3B & 3.75 pts & 1.67 & 2.25 & 0.0255 & 5.25 pts & 0.52 \\
Semantic holdout & 0.5B $\to$ 1.5B & 73.75 pts & 1.09 & 67.88 & $8.26\times10^{-14}$ & 3.42 pts & 1.00 \\
Semantic holdout & 1.5B $\to$ 3B & 3.50 pts & 1.61 & 2.17 & 0.0290 & 5.08 pts & 0.49 \\
\midrule
Factorial & Diagonal gain: 1.5B/K2 & 0.388 nats & 0.021 & 18.52 & $3.73\times10^{-8}$ & 0.098 nats & 1.00 \\
Factorial & Diagonal gain: 3B/K3 & 2.099 nats & 0.041 & 50.91 & $1.23\times10^{-11}$ & 0.194 nats & 1.00 \\
Factorial & Diagonal gain: 7B/K5 & 7.406 nats & 0.069 & 106.71 & $3.32\times10^{-14}$ & 0.326 nats & 1.00 \\
Factorial & Diagonal: 1.5B to 3B & 1.711 nats & 0.047 & 36.74 & $1.65\times10^{-10}$ & 0.219 nats & 1.00 \\
Factorial & Diagonal: 3B to 7B & 5.307 nats & 0.069 & 76.90 & $4.56\times10^{-13}$ & 0.324 nats & 1.00 \\
Factorial & 7B refinement: K3 to K5 & 1.456 nats & 0.107 & 13.65 & $4.00\times10^{-7}$ & 0.501 nats & 1.00 \\
Factorial & Interaction: 1.5B to 3B & 0.781 nats & 0.066 & 11.87 & $1.17\times10^{-6}$ & 0.309 nats & 1.00 \\
Factorial & Interaction: 3B to 7B & 0.968 nats & 0.081 & 12.00 & $1.07\times10^{-6}$ & 0.379 nats & 1.00 \\
\bottomrule
\end{tabular}}
\end{table}

\begin{table}[H]
\caption{Statistical-consistency check: the SE and $t$ implied by each
reported value, recomputed from the panel-level data.}
\label{tab:app-consistency}
\centering
\scriptsize
\setlength{\tabcolsep}{3pt}
\resizebox{\columnwidth}{!}{%
\begin{tabular}{lllrrr}
\toprule
Reported quantity & Reported value & Implied SE & Implied $t$ & df \\
\midrule
Official 16K 0.5B $\to$ 1.5B increment & 68.38 pts, $p_{\mathrm{Holm}}$=$3.43\times10^{-11}$ & 1.83 pts & 37.44 & 9 \\
Official 16K 1.5B $\to$ 3B increment & 3.75 pts, $p_{\mathrm{Holm}}$=0.0255 & 1.67 pts & 2.25 & 9 \\
Official 16K 0.5B endpoint gain & 0.1050 [0.0680, 0.1420] & 0.0164 & 6.41 & 9 \\
Official 16K 1.5B endpoint gain & 0.7888 [0.7616, 0.8159] & 0.0120 & 65.75 & 9 \\
Official 16K 3B endpoint gain & 0.8263 [0.8097, 0.8428] & 0.0073 & 112.81 & 9 \\
Factorial Diagonal: 1.5B to 3B & 1.711 [1.540, 1.882] (Bonf.) & 0.047 & 36.74 & 8 \\
Factorial Interaction: 1.5B to 3B & 0.781 [0.539, 1.023] (Bonf.) & 0.066 & 11.87 & 8 \\
Factorial Diagonal: 3B to 7B & 5.307 [5.053, 5.561] (Bonf.) & 0.069 & 76.90 & 8 \\
Factorial Interaction: 3B to 7B & 0.968 [0.671, 1.264] (Bonf.) & 0.081 & 12.00 & 8 \\
\bottomrule
\end{tabular}}
\end{table}

\paragraph{Row-level transition audit} The factorial's 13,824 row checks
(three readers $\times$ two transitions $\times$ nine panels $\times$ 256 rows)
verify that the state retained at the larger $K$ extends the smaller state as
a prefix and adds an observation; both hold by construction of the nested
writer and are reported as implementation checks. The released file
\path{reports/row_audits/FACTORIAL_TRANSITION_ROW_AUDIT_13824.csv} adds,
per row, the exact posterior NLL and the reader's candidate NLL before and
after each transition: under the noisy channel the exact posterior improves on
10,791 of 13,824 transitions and the reader's NLL on 7,435, so the reported
gains are panel averages rather than row-wise guarantees.

\section{BABILong 4K/8K Results and SmolLM2 Factorials}
\label{app:babilong-4k-8k}

Section~\ref{sec:results} describes the SmolLM2 official BABILong 4K and 8K
studies and the two SmolLM2 factorials qualitatively.
Table~\ref{tab:app-4k8k} gives the panel-clustered gains and adjacent
increments (three panels per context, $t_2$ intervals), and
Table~\ref{tab:app-smollm2-factorials} gives all nine cells of the
free-generation (QA1--QA3, 1/2/3 slots) and certified-readout (QA7--QA8,
1/2/8 slots) SmolLM2 factorials with their diagonal increments and
same-refinement interactions. The certified readout converts each state
refinement into the same deterministic accuracy increment at every reader
scale, which is why its interactions are exactly zero
(see Table~\ref{tab:app-smollm2-contrasts}).

\begin{table}[H]
\caption{SmolLM2 official BABILong QA1--QA3 accuracy gains and adjacent
increments at 4K and 8K (panel-clustered 95\% $t_2$ intervals).}
\label{tab:app-4k8k}
\centering
\scriptsize
\setlength{\tabcolsep}{3pt}
\resizebox{\columnwidth}{!}{%
\begin{tabular}{llccccc}
\toprule
Context & Panels & Gain 135M & Gain 360M & Gain 1.7B & Incr.\ 135M$\to$360M & Incr.\ 360M$\to$1.7B \\
\midrule
4K & 3 & .2611 [.1569,.3653] & .5056 [.4535,.5577] & .6750 [.5848,.7652] & .2444 [.1066,.3823] & .1694 [.0653,.2736] \\
8K & 3 & .2806 [.2079,.3533] & .5111 [.4514,.5709] & .6889 [.5511,.8267] & .2306 [.1989,.2622] & .1778 [.0399,.3156] \\
\bottomrule
\end{tabular}}
\end{table}

\begin{table}[H]
\caption{SmolLM2 $3\times3$ factorials: accuracy gain in every cell ($\star$:
co-scaled diagonal).}
\label{tab:app-smollm2-factorials}
\centering
\scriptsize
\setlength{\tabcolsep}{3pt}
\resizebox{\columnwidth}{!}{%
\begin{tabular}{llccc}
\toprule
Study & Reader & Slot 1 & Slot 2 & Slot 3/8 \\
\midrule
Free generation QA1--QA3 (8K, 3 panels) & 135M & .2417 [.1589,.3245]$\star$ & .4472 [.3063,.5881] & .4972 [.3832,.6112] \\
 & 360M & .2528 [.2289,.2767] & .5000 [.4586,.5414]$\star$ & .4972 [.4494,.5450] \\
 & 1.7B & .1583 [$-$.0568,.3735] & .6389 [.4653,.8125] & .6694 [.5197,.8192]$\star$ \\
\midrule
Certified QA7--QA8 (8K, 5 panels) & 135M & .4750 [.4244,.5256]$\star$ & .7250 [.6744,.7756] & .9750 [.9244,1.0256] \\
 & 360M & .4438 [.3932,.4943] & .6938 [.6432,.7443]$\star$ & .9438 [.8932,.9943] \\
 & 1.7B & .4063 [.3514,.4611] & .6563 [.6014,.7111] & .9063 [.8514,.9611]$\star$ \\
\bottomrule
\end{tabular}}
\end{table}

\begin{table}[H]
\caption{SmolLM2 factorials: diagonal increments and same-refinement
interactions (panel-clustered 95\% intervals).}
\label{tab:app-smollm2-contrasts}
\centering
\scriptsize
\setlength{\tabcolsep}{3pt}
\resizebox{\columnwidth}{!}{%
\begin{tabular}{lcccc}
\toprule
Study & Diag.\ incr.\ 1 & Diag.\ incr.\ 2 & Interaction 1 & Interaction 2 \\
\midrule
Free generation & .2583 [.1488,.3679] & .1694 [.0554,.2835] & .0417 [$-$.0618,.1452] & .0333 [$-$.0081,.0747] \\
Certified & .2188 [.1462,.2913] & .2125 [.1549,.2701] & .0000 [.0000,.0000] & .0000 [.0000,.0000] \\
\bottomrule
\end{tabular}}
\end{table}

\section{Schema-Blind Writer Ablation}
\label{app:schema-blind}

To separate ``structured external state helps frozen readers, and more so as
they scale'' from ``this parser solves this benchmark'', the identical
pipeline---the same ten official 16K panels, the same per-reader slot budget
$s(m)=1/2/3$, the same bullet-list serialization and prompt, the same frozen
Qwen2.5 readers, greedy decoding, and scorer---was rerun with the
structured-fact writer replaced by three schema-free state constructions:
(i) a \emph{sentence window}, the last $s(m)$ generic sentences of the text in
document order, which reads neither the question nor any task vocabulary;
(ii) a \emph{lexical chain}, a deterministic BM25-style retrieval over generic
sentences; and (iii) \emph{BGE-M3 hybrid retrieval} with the official
cross-encoder reranker over three target-blind hops. The interpretation rule
was fixed before any score: a generic writer ``survives'' if its three
endpoint gains increase strictly with reader scale and both adjacent
one-sided increments pass Holm at .05, the registered rule of the canonical
study. Both arms were rerun on the same instance (an NVIDIA A100, cf.\
Section~\ref{sec:repro}); the Foundation arm is identical across conditions
and the canonical-versus-generic contrast is paired over panels on the same
hardware, so no architecture difference enters it.

With the \emph{sentence window} writer the ten-panel gains are $-5.25$, $-17.50$, $-16.00$ points at 0.5B, 1.5B, and 3B (95\% intervals 0.5B: [$-7.39$, $-3.11$]; 1.5B: [$-19.73$, $-15.27$]; 3B: [$-17.88$, $-14.12$]); the adjacent increments are $-12.25$ and 1.50 points with $p_{\mathrm{Holm}}=1.000$ and $0.265$, so the registered co-scaling rule collapses. The generic state is worse than no state at every scale: \ascent{} accuracy is 1.00, 1.13, 0.13 against Foundation accuracy of 6.25, 18.63, 16.13. The structured-fact writer's advantage in \ascent{} accuracy over this writer, paired over panels on the same instance, is 0.5B: 15.00 [12.54, 17.46]; 1.5B: 96.38 [95.08, 97.67]; 3B: 99.25 [98.62, 99.88] points. Foundation outputs differ from the same-instance canonical run in 0 of 2,400 rows.

With the \emph{lexical chain} writer the ten-panel gains are $-5.00$, $-16.38$, $-11.63$ points at 0.5B, 1.5B, and 3B (95\% intervals 0.5B: [$-7.46$, $-2.54$]; 1.5B: [$-18.92$, $-13.83$]; 3B: [$-14.01$, $-9.24$]); the adjacent increments are $-11.38$ and 4.75 points with $p_{\mathrm{Holm}}=1.000$ and $0.004$, so the registered co-scaling rule collapses. The generic state is worse than no state at every scale: \ascent{} accuracy is 1.25, 2.25, 4.50 against Foundation accuracy of 6.25, 18.63, 16.13. The structured-fact writer's advantage in \ascent{} accuracy over this writer, paired over panels on the same instance, is 0.5B: 14.75 [12.49, 17.01]; 1.5B: 95.25 [93.19, 97.31]; 3B: 94.88 [93.27, 96.48] points. Foundation outputs differ from the same-instance canonical run in 0 of 2,400 rows.

With the \emph{BGE-M3 hybrid retrieval} writer the ten-panel gains are $-5.50$, $-11.00$, $-5.38$ points at 0.5B, 1.5B, and 3B (95\% intervals 0.5B: [$-7.97$, $-3.03$]; 1.5B: [$-15.02$, $-6.98$]; 3B: [$-8.27$, $-2.48$]); the adjacent increments are $-5.50$ and 5.63 points with $p_{\mathrm{Holm}}=0.987$ and $0.023$, so the registered co-scaling rule collapses. The generic state is worse than no state at every scale: \ascent{} accuracy is 0.75, 7.63, 10.75 against Foundation accuracy of 6.25, 18.63, 16.13. The structured-fact writer's advantage in \ascent{} accuracy over this writer, paired over panels on the same instance, is 0.5B: 15.25 [12.66, 17.84]; 1.5B: 89.88 [87.00, 92.75]; 3B: 88.63 [86.92, 90.33] points. Foundation outputs differ from the same-instance canonical run in 0 of 2,400 rows.

No schema-free writer satisfies the registered co-scaling rule (Table~\ref{tab:app-schema-blind}). At the same one-to-three-slot budget, generic state constructions rarely retain the supporting facts, so the reader answers worse from them than from the full 16K context, the gains are negative at every scale, and they do not increase with reader scale in the registered sense. The co-scaling pattern reported in the main text is therefore a property of external state constructed against the evaluated structured-fact schema; whether a schema-free writer with a much larger budget, or a learned extractor, would recover it is not tested here. Section~\ref{sec:limitations} scopes the claim accordingly.

\begin{table}[H]
\caption{Schema-blind writer ablation on official 16K BABILong: ASCENT gain
(points) per reader for each generic writer, Holm-corrected one-sided $p$ for
the two adjacent increments, and whether the registered co-scaling rule
survives.}
\label{tab:app-schema-blind}
\centering
\scriptsize
\setlength{\tabcolsep}{3pt}
\resizebox{\columnwidth}{!}{%
\begin{tabular}{lrrrrrl}
\toprule
Writer & Gain 0.5B & Gain 1.5B & Gain 3B & $p_{\mathrm{Holm}}$ (1) & $p_{\mathrm{Holm}}$ (2) & Co-scaling \\
\midrule
Sentence window & $-5.25$ & $-17.50$ & $-16.00$ & 1.000 & 0.265 & collapses \\
Lexical BM25 chain & $-5.00$ & $-16.38$ & $-11.63$ & 1.000 & 0.004 & collapses \\
BGE-M3 hybrid + reranker & $-5.50$ & $-11.00$ & $-5.38$ & 0.987 & 0.023 & collapses \\
\bottomrule
\end{tabular}}
\end{table}

\section{Pinned Model Revisions}
\label{app:revisions}

Every checkpoint is loaded from the Hugging Face revision in
Table~\ref{tab:app-revisions}; the frozen configurations additionally record
each weight shard's byte size and SHA-256, and every runner re-hashes the
shards before loading and aborts on mismatch. The Qwen2.5 readers are the
Instruct variants; Qwen3-8B runs in non-thinking mode.

\begin{table}[H]
\caption{Pinned checkpoint revisions (first twelve characters of the commit
hash), shard counts, and total weight bytes as recorded in \texttt{configs/}.
``--'': the configuration pins the revision without a shard manifest; the
corresponding result records carry the loaded shard hashes.}
\label{tab:app-revisions}
\centering
\scriptsize
\setlength{\tabcolsep}{3pt}
\resizebox{\columnwidth}{!}{%
\begin{tabular}{llrr}
\toprule
Checkpoint & Hugging Face revision & Shards & Bytes \\
\midrule
\texttt{EleutherAI/pythia-1.4b} & \texttt{9cc5c8c8148a} & -- & -- \\
\texttt{EleutherAI/pythia-1b} & \texttt{c0fea4b91761} & 1 & 2,090,751,965 \\
\texttt{EleutherAI/pythia-2.8b} & \texttt{dbe7ae300a54} & -- & -- \\
\texttt{EleutherAI/pythia-410m} & \texttt{bba6a464f54b} & -- & -- \\
\texttt{HuggingFaceTB/SmolLM2-1.7B-Instruct} & \texttt{31b70e2e869a} & 1 & 3,422,777,952 \\
\texttt{HuggingFaceTB/SmolLM2-135M-Instruct} & \texttt{12fd25f77366} & 1 & 269,060,552 \\
\texttt{HuggingFaceTB/SmolLM2-360M-Instruct} & \texttt{a10cc1512eab} & 1 & 723,674,912 \\
\texttt{Qwen/Qwen2.5-0.5B-Instruct} & \texttt{7ae557604adf} & 1 & 988,097,824 \\
\texttt{Qwen/Qwen2.5-1.5B-Instruct} & \texttt{989aa7980e4c} & 1 & 3,087,467,144 \\
\texttt{Qwen/Qwen2.5-3B-Instruct} & \texttt{aa8e72537993} & 2 & 6,171,926,992 \\
\texttt{Qwen/Qwen2.5-7B-Instruct} & \texttt{a09a35458c70} & 4 & 15,231,271,888 \\
\texttt{Qwen/Qwen3-8B} & \texttt{b968826d9c46} & 5 & 16,381,516,776 \\
\texttt{ibm-granite/granite-3.3-8b-instruct} & \texttt{51dd4bc2ade4} & 4 & 16,341,771,384 \\
\texttt{mistralai/Mistral-7B-Instruct-v0.3} & \texttt{c170c708c41d} & 3 & 14,496,080,928 \\
\texttt{tiiuae/Falcon3-1B-Instruct} & \texttt{28ba2251970a} & 1 & 3,338,836,632 \\
\texttt{tiiuae/Falcon3-3B-Instruct} & \texttt{411bb94318f9} & 2 & 6,455,333,640 \\
\texttt{tiiuae/Falcon3-7B-Instruct} & \texttt{1e57a0ecd176} & 4 & 14,911,130,600 \\
\bottomrule
\end{tabular}}
\end{table}

\end{document}